# DEEP CONVOLUTIONAL NEURAL NETWORK APPLICATION ON ROOFTOP DETECTION FOR AERIAL IMAGE

*Mengge Chen[1], Jonathan Li[1*2]*

[1] Department of Geography and Environmental Management
[2] Department of Systems Design Engineering
University of Waterloo, 200 University Avenue West, Waterloo, Ontario N2L 3G1, Canada
*Corresponding author: junli@uwaterloo.ca

## ABSTRACT

As one of the most destructive disasters in the world, earthquake causes death, injuries, destruction and enormous damage to the affected area. It is significant to detect buildings after an earthquake in response to reconstruction and damage evaluation. In this research, we proposed an automatic rooftop detection method based on the convolutional neural network (CNN) to extract buildings in the city of Christchurch and tuned hyperparameters to detect small detached houses from the aerial image. The experiment result shows that our approach can effectively and accurately detect and segment buildings and has competitive performance.

## I. INTRODUCTION

Earthquake is one of the most catastrophic natural disasters that threaten the human being. It accounts for 36% of the average annual losses from natural hazards, and the building collapse after the earthquake directly causes casualties and property losses [1]. In the Canterbury area in New Zealand, there are many seismic events and major earthquakes that caused dwelling and infrastructure damage since 2010 [2]. The disaster resulted in substantial land damage across the Christchurch city, as well as the dramatic modification of the built and natural environment. Therefore, the assessment of buildings in the urban area becomes an important task in the reconstruction and recovery phases.

The automatic detection of the rooftop is significant for severe building damage statistics after a destructive disaster event like earthquakes. The remote sensing technology can be recognized as an effective and low-risk method to obtain timely data about automatic detection of damaged buildings, especially for large areas [3]. The image-based rooftop detection with very high spatial resolution using the unmanned aerial vehicle (UAV) therefore can show the detail view of the damaged building and achieve high accuracy [4].

Various algorithms have been proposed through the literature that detect and extract buildings from the remote sensing images. Traditional building detection and segmentation methods use training samples to predict synonymous features, such as the Maximum Likelihood Classifier (MLC) and Support Vector Machine (SVM) and Random Forest (RF) [5-7]. However, these methods can only accomplish an acceptable performance due to the increased data complexity and dimensionality which is difficult to use simple features to predict the class label [6]. Recently, feature extraction using the deep convolutional neural network (DCNN) outperforms the previous methods in remote sensing processing, including the semantic segmentation, object detection and scene classification [5]. [8] proposed a convolution neural network (CNN) model, combined with the multi-labeling layer (ML) composed of a customized thresholding operation, to classify multi-labeled UAV images. [9] applied a different network architecture in deep learning to enhance the building segmentation accuracy. The paper used fully convolutional networks (FCNs) to segment buildings and roads from the high spatial resolution imagery [9]. Furthermore, Xu et al. [10] improved the building classification result from Very High Resolution (VHR) imagery by a deep learning model based on deep residual networks (ResNet), which is specified as Res-U-Net, and added a guided filter to optimize the extraction buildings in the post-processing section. In order to enhance the building extraction from UAV image, [11] proposed a building segmentation algorithm based on building boundary constraints using oblique UAV images and optimized the building footprint generation from OpenStreetMap (OSM).

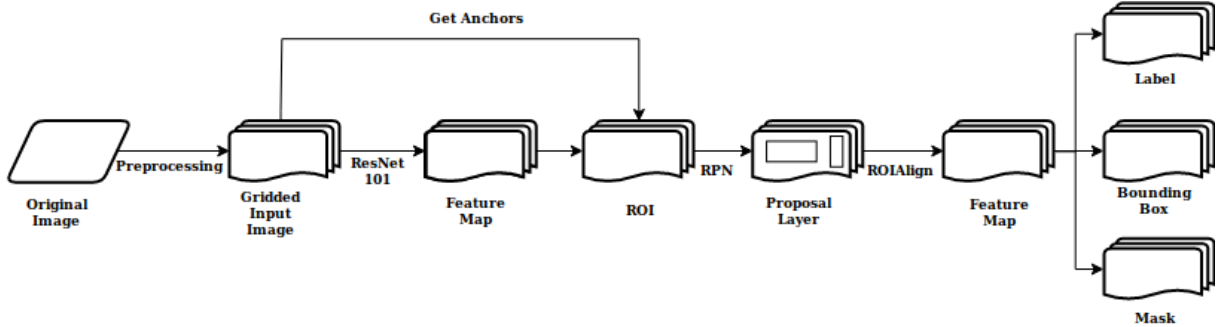

Fig. 1. Flowchart of the method

The objective of this paper is to implement an automatic building detection method based on the state-of-the-art deep learning method. CNN will be used as our model to detect the rooftop in the Christchurch Post-Earthquake Urban Aerial dataset, and the detection result will be compared to a pre-trained deep learning model in terms of the accuracy assessment.

## II. METHODOLOGY

Fig.1 illustrates the workflow of our research based on Mask R-CNN including three main stages. The first stage is the feature extraction using an existing deep learning model, and the second stage is the Regional Proposal Network (RPN), with a binary mask generation and object classification as the last stage. The following section will explain the proposed framework in detail.

Before applying the Mask R-CNN on the image, the aerial dataset requires a preprocessing that uses a grid to divide the whole image into tiles of patches. Mask R-CNN, a general framework for object detection and segmentation, is used as the baseline of our method. The gridded images are then passed through a convolutional neural network to obtain the feature map. In this research, we choose ResNet 101 as the basic network architecture for feature extraction.

After the CNN computed, anchors are selected uniformly across the feature map, and for each anchor, nine different sizes of the region of interests (RoIs) are chosen and connected to the original image input. The RPN is then applied to these candidate RoIs to filter the irrelevant bounding boxes (BB) by object & background classification and BB regression. The classification in RPN is a binary classification to classify the content in the BB into two labels: the object or the background label. The BB regression aims to modify the BB coordinates to better fit the objects.

RoIAlign takes an important role in the feature map extraction from each RoI and improves the extraction performance compared to RoIPool in Faster R-CNN [12]. By using bilinear interpolation to calculate the exact pixel value of the object surrounded by four real sampled pixels, RoIAlign removes the quantitative operation and achieves higher detection accuracy. Therefore, the feature map with the fixed size is exported to the Fast R-CNN classifier where SoftMax probability estimation is applied on the fully-connected layer to produce the final result for classification labels (one is building, and one is background in this research) and BB locations. Additionally, the feature map is fed into the mask segmentation branch to derive the building mask using a small FCN.

## III. EXPERIMENT

### A. Dataset

The experimental dataset is aerial photos collected from the Land Information New Zealand on April 2014 in response to the 22 February 2011 earthquake in Canterbury [13]. The ortho-rectified RGB images are in the New Zealand Transverse Mercator (NZTM) projection with a ground resolution of 0.1m. The aerial imagery covers 450 km2 of city Christchurch, New Zealand, including more than 220, 000 independent buildings. The ground truth data or mask data is manually edited by the laboratory from Wuhan University, which is a binary image that represents the building and non-building pixels [14]. We split the original image into 512*512-pixel sub-images; the whole dataset was then divided into the training dataset (80%) and testing dataset (20%).

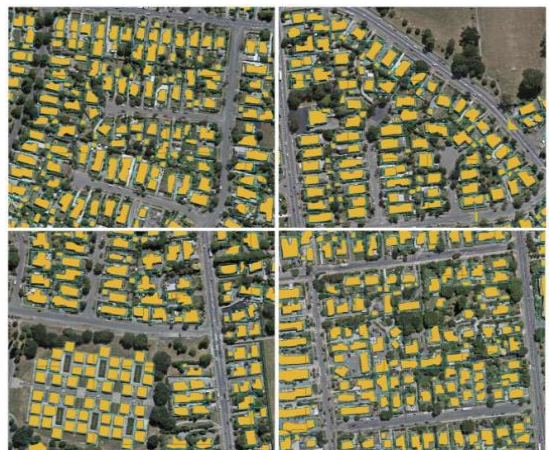

Fig. 2. Examples of the rooftop detection result

*B. Qualitative and Quantitative Results*

As can be seen from Fig. 2, the rooftop of detached houses in city Christchurch are automatically and correctly detected and extracted: the yellow masks are surrounded by blue bounding box and cover most of the buildings in the map. Although the output requires post-processing to fit the mask to the actual building edges, the overall result shows the robustness and efficiency of the deep learning method on the extraction of building rooftop.

Table 1. Model performance of the proposed method and pretrained Mask R-CNN

| Method | Precision | Recall | F1-Score |
| --- | --- | --- | --- |
| Proposed Method | 75.0% | 83.5% | 79.0% |
| Pretrained Mask R-CNN | 69.2% | 80.3% | 74.3% |

The accuracy assessment (Table 1) was conducted as well to explain the result quantitatively, and the proposed model is compared to a pre-trained model from the Mask R-CNN baseline solution [15]. The Intersection over Union (IoU) threshold is set to 0.5, meaning that all the predicted masks with an overlap over 50% with the ground truth data will be considered as a successful detection. The true positive (TP), true negative (FP), false negative (FN) are calculated to obtain the precision and recall statistic; and the F1-Score that combines the precision and recall value is presented as well. According to the accuracy assessment, our method has better detection performance, and the F1-Score of 79% represents the reasonable balance between precision and recall.

## IV. CONCLUSION

In this paper, a rooftop detection method is proposed based on the advanced deep learning algorithm. The Mask R-CNN is implemented as the framework of the detection method; specifically, the ResNet is chosen as the network backbone; then the RPN is applied on the feature maps, and RoIAlign is used for feature match. The objects in the image are classified into two types: the background or the building, following a mask branch that segments building rooftop from the background. Experiments show that the proposed detection model is efficient and feasible to extract a detached house from aerial images, which is meaningful for the post-earthquake recovery and damage evaluation. For feature works, more training data is required to provide a more accurate detection result, and the building edge issue is to be noted in the post-processing stage.